\documentclass[a4paper,twoside]{article}
\usepackage{epsfig}
\usepackage{subcaption}
\usepackage{calc}
\usepackage{amssymb}
\usepackage{amstext}
\usepackage{amsmath}
\usepackage{amsthm}
\usepackage{multicol}
\usepackage{pslatex}
\usepackage{algorithm2e}
\usepackage[bottom]{footmisc}
\usepackage{booktabs} 
\usepackage{graphicx}
\usepackage{url}
\usepackage{SCITEPRESS}
\usepackage{natbib}
\usepackage{verbatim}

\newcommand\blfootnote[1]{%
  \begingroup
  \renewcommand\thefootnote{}\footnote{#1}%
  \addtocounter{footnote}{-1}%
  \endgroup
}

\begin{document}

\title{How to Sample High Quality 3D Fractals for Action Recognition Pre-Training?}

\author{\authorname{Marko Putak\sup{1}\orcidAuthor{0009-0001-8709-9765}, Thomas B. Moeslund\sup{1,2}\orcidAuthor{0000-0001-7584-5209}, and Joakim Bruslund Haurum\sup{2,3}\orcidAuthor{0000-0002-0544-0422}}
\affiliation{\sup{1}Visual Analysis and Perception Lab, Aalborg University, Aalborg, Denmark}
\affiliation{\sup{2}Pioneer Centre for AI, Denmark, \sup{3}Center for Software Technology, University of Southern Denmark, Vejle, Denmark}
\email{\{mapu, tbm\}@create.aau.dk, jhau@mmmi.sdu.dk}}


\abstract{Synthetic datasets are being recognized in the deep learning realm as a valuable alternative to exhaustively labeled real data. One such synthetic data generation method is Formula Driven Supervised Learning (FDSL), which can provide an infinite number of perfectly labeled data through a formula driven approach, such as fractals or contours. FDSL does not have common drawbacks like manual labor, privacy and other ethical concerns. In this work we generate 3D fractals using 3D Iterated Function Systems (IFS) for pre-training an action recognition model. The fractals are temporally transformed to form a video that is used as a pre-training dataset for downstream task of action recognition. We find that standard methods of generating fractals are slow and produce degenerate 3D fractals. Therefore, we systematically explore alternative ways of generating fractals and finds that overly-restrictive approaches, while generating aesthetically pleasing fractals, are detrimental for downstream task performance. We propose a novel method, Targeted Smart Filtering, to address both the generation speed and fractal diversity issue. The method reports roughly $100$ times faster sampling speed and achieves superior downstream performance against other 3D fractal filtering methods.}

\onecolumn \maketitle \normalsize \setcounter{footnote}{0} \vfill

\section{\uppercase{Introduction}}
\label{sec:introduction}

\blfootnote{Code and datasets are available at: \\ 
\hspace*{1em}\url{https://github.com/mputak/3d-fractal-tsf}}

In recent years, deep learning has become the dominant paradigm for video action recognition. However, the success of these data-hungry models is dependent on the availability of large-scale, often annotated video datasets. The collection and annotation process of such datasets are filled with challenges: they are resource-intensive, expensive and introduce significant privacy and ethical concerns (e.g., GDPR compliance). Synthetic datasets have emerged as a powerful alternative to these problems. Specifically, Formula Driven Supervised Learning (FDSL) \citep{KataokaIJCV2022, KataokaACCV2020} stands out as a promising approach to generating virtually infinite, perfectly labeled datasets without any manual annotation labor. The potential for infinite data is rooted in the uncountably infinite set of real-valued parameters that define the generating formulas, and is practically realized by the dynamic, stochastic nature of the generation process. By using mathematical formulas, such as Iterated Function Systems (IFS) for fractals, it is possible to auto-generate complex visual data and use the generation parameters as intrinsic labels. This intrinsic labeling ensures perfect fidelity, as the label is defined by the exact parameters used to construct the data instance, eliminating all human error or subjective visual inspection associated with manual annotation.

\begin{figure*}[!t]
    \centering
    \includegraphics[width=0.95\linewidth]{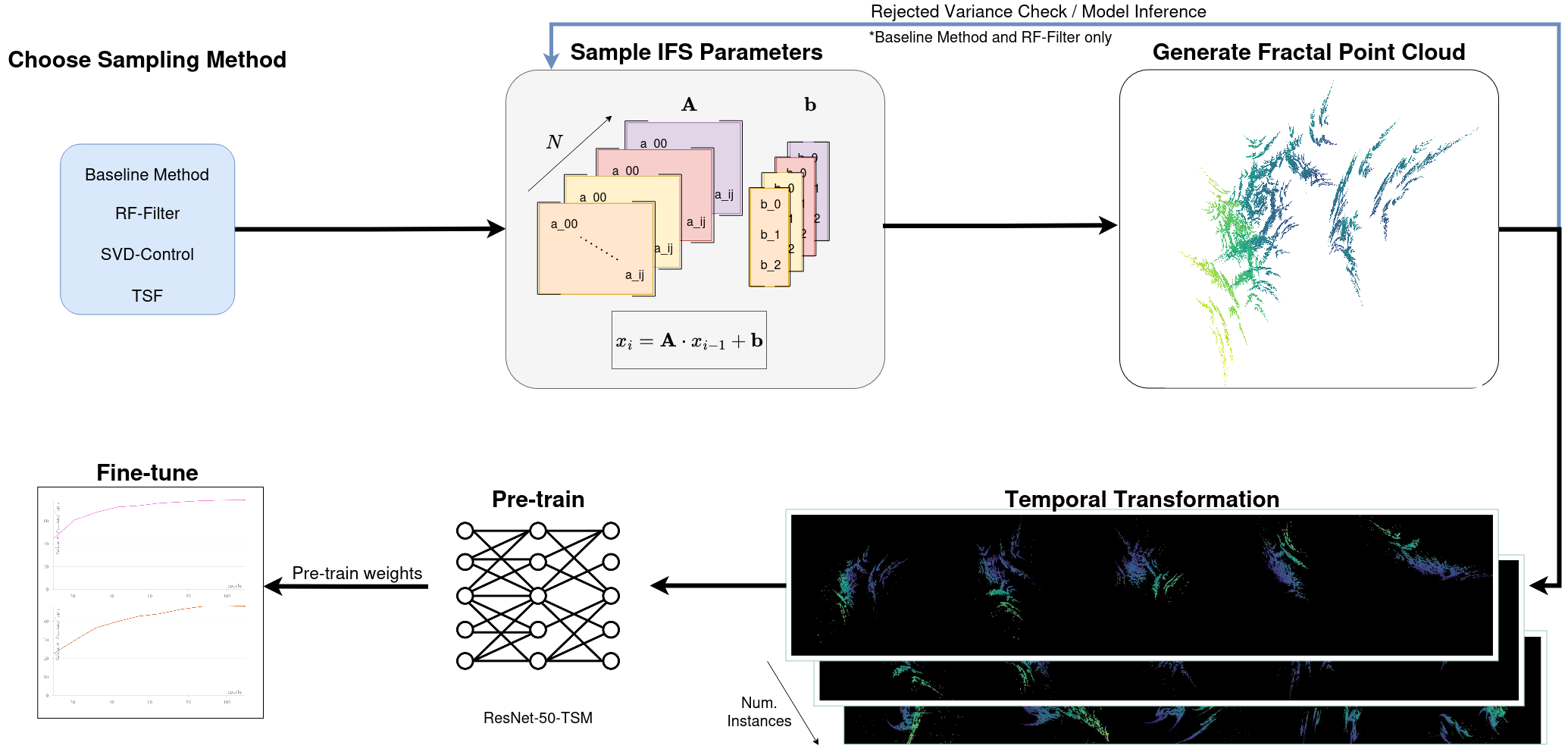}
    \caption{The Complete 3D Fractal Video Pre-Training Pipeline. The process begins with finding valid IFS parameters via one of four main methods. These parameters generate a 3D fractal point cloud using the Chaos Game, which is dynamically transformed into a video dataset. The final dataset is used for model pre-training, and the resulting weights are fine-tuned and compared against baselines on downstream action recognition tasks.}
    \label{fig:lead_fig}
\end{figure*}
While FDSL has shown promise, most work in this area has focused on 2D images or 2D fractal videos \citep{KataokaACCV2020, improving, visual_atoms, 2dfractals_ar}, with little attention to 3D fractals \citep{mv-fractaldb}. 3D fractals, generated as point clouds, offer inherently richer spatial and structural complexity. We therefore hypothesize that pre-training on dynamic 3D structures would better instill powerful spatio-temporal representations beneficial for real-world downstream video tasks.

However, generating 3D fractals is non-trivial. Naive parameter sampling often produces degenerate geometries. Some point sets are overly contractive, collapsing into a line or plane, while others become excessively sparse and lack structure. In the 3D case, each transformation matrix is $3\times3$ rather than $2\times2$ as in 2D, which exponentially expands the parameter space and increases the likelihood of degeneracy.

This leads to our central research question: \textit{How can we efficiently filter the vast 3D IFS parameter space to remove degenerate samples while preserving the geometric diversity crucial for learning transferable features?} Intuitively, one might assume that a dataset consisting of only geometrically and aesthetically complex fractals would yield superior transferable features.

We propose a full pipeline to generate 3D fractal videos by applying spatio-temportal transformations to 3D fractal point clouds generated from Iterated Function Systems (IFS). The overall structure of our proposed 3D Fractal FDSL pipeline is illustrated in Figure \ref{fig:lead_fig}. The generated datasets are filtered using four distinct generation/filtering methods, namely Naive sampling with variance filtering, Singular Value Decomposition (SVD) filtering, binary classification filtering, and our proposed Target Smart Filtering method. These methods are systematically compared in terms of downstream performance on the UCF101 \citep{ucf101} and HMDB51 \citep{hmdb51} action recognition datasets, as well as sampling efficiency. 

Our contributions are as follows:

\begin{enumerate}
    \item We find that pre-training on videos of synthetic 3D fractals improves performance significantly over training from scratch on both UCF101 and HMDB51,
    \item The SVD-Control and Data-Driven Random Forest Filtering methods are overly restrictive, performing worse than the Naive + Variance Filtering baseline.
    \item Our proposed Targeted Smart Filtering method outperforms or matches that of the three prior considered filtering methods, while generating samples $100$ times faster than the Naive + Variance filtering baseline.
\end{enumerate}

\section{\uppercase{Related Work}}

This research is situated at the intersection of three key domains: 1) pre-training strategies for video, 2) synthetic data generation, and 3) the specific sub-field of fractal-based learning. This overview will cover the foundational work in each area to contextualize novel contributions to 3D fractal generation for action recognition.

\subsection{Pre-Training for Video Action Recognition}

The dominant pre-training paradigm for action recognition is to utilize supervised learning on a massive, real-world dataset such as the Kinetics dataset \citep{kinetics}. Models pre-trained on Kinetics, including the I3D model by \citet{i3d}, have become the de facto baseline of the field. However, the immense cost of data collection and human annotation, data-licensing issues and privacy concerns hinder the usability of the dataset. 

Concurrently, there has been a significant evolution of the underlying neural network architecture. Starting from the early 2D-CNN coupled with LSTM from \citet{2dcnnrnn} and 3D CNN by \citet{3dcnn} which are computationally expensive, to more efficient two-stream networks \citep{twostream}. This led to the development of efficient temporal modeling through the Temporal Shift Module (TSM) introduced by \citet{tsm}. This module achieves high performance by enabling temporal reasoning in a 2D CNN backbone with almost zero computational overhead.

For a study based on data generation, where many training runs are expected, the ResNet-50-TSM architecture represents the ideal baseline, as it is strong, efficient and widely recognized architecture. Finally, it also allows for a fair comparison with prior work such as \citet{2dfractals_ar}.

\subsection{Synthetic Data and Formula-Driven Supervised Learning (FDSL)}

Synthetic data generation in the computer vision field has gained increasing attention in recent years. The types of synthetic data and their applications span a broad spectrum, including datasets for autonomous driving such as CARLA \citep{carla}, procedurally generated patterns like Perlin noise \citep{perlin}, and generative models such as GANs \citep{gans, zhang21}. 

Formula-Driven Supervised Learning (FDSL) \citep{fdsl}, a recently introduced sub-field of synthetic data generation, is known for its exploitation of mathematical formulas to generate paired data and labels. FDSL was originally proposed by \citet{KataokaACCV2020}, who demonstrated that strong transfer learning performance can be achieved without using natural images. Their groundbreaking idea was to create FractalDB, a large dataset of 2D fractal images generated via Iterated Function Systems, and to use the generation parameters as class labels, serving as a substitute for pre-training on ImageNet \citep{Russakovsky2015}. 
Alternative data generation approaches have since been considered for FDSL, such as Visual Atom by \citet{visual_atoms} who used concentric sinusoidal wave patterns instead of fractals. This achieved near real image dataset performance with $14$ times smaller dataset, validating that structured, formula-driven data is a powerful tool for pre-training.

\subsection{The Evolution of Fractal Generation for Pre-Training}

The 2D foundation for fractal pre-training was introduced by \cite{KataokaACCV2020} showcasing the pre-training capabilities of fractals. An improvement to the naive sampling used in the original formulation was proposed by \citet{improving}. They observed that the naive sampling produced many degenerate (e.g., too contractive) 2D fractals. Instead, they proposed an intricate solution that utilized sampling parameters of singular value decomposed matrix $\mathbf{A}$ instead of the direct matrix entries. This allowed them to guarantee contractivity by constraining the singular values ($\sigma$-factor). From there, they heuristically found a rule to generate quality fractals with system size and singular values as input variables, by fitting a Linear SVM on the $\sigma$-factors to classify good and bad fractals. Their research established State-of-the-Art for 2D fractal quality filtering.

In terms of 3D fractals, \citet{mv-fractaldb} introduced 3D fractal point clouds for 3D object recognition. Crucially, they used naive uniform sampling to directly find matrix $A$ and vector $b$ elements, followed by a simple post-hoc variance filter to remove degenerate point clouds.

Finally, \citet{2dfractals_ar} used 2D fractals to generate videos by interpolating between two different 2D fractal images and applying other transformations for intra-class diversity. This work is the first to prove that fractal videos are highly effective for pre-training action recognition.

\subsection{Our Contribution in Context}

Our work builds directly on these foundations to address clear, un-answered questions at the intersection of these fields. Firstly, while \citet{2dfractals_ar} demonstrated the efficacy of 2D fractal videos and \citet{mv-fractaldb} used static 3D point clouds, the generation and utility of dynamic 3D fractal videos for action recognition pre-training remains an unexplored area. 
Secondly, while there has been proposed improvements for sampling fractals in 2D \citep{improving}, no such evaluation has been carried out for 3D fractals. We aim to rectify this in this work.

\begin{figure*}
    \centering
    \begin{subfigure}{0.5\textwidth}
        \centering
        \includegraphics[width=\linewidth]{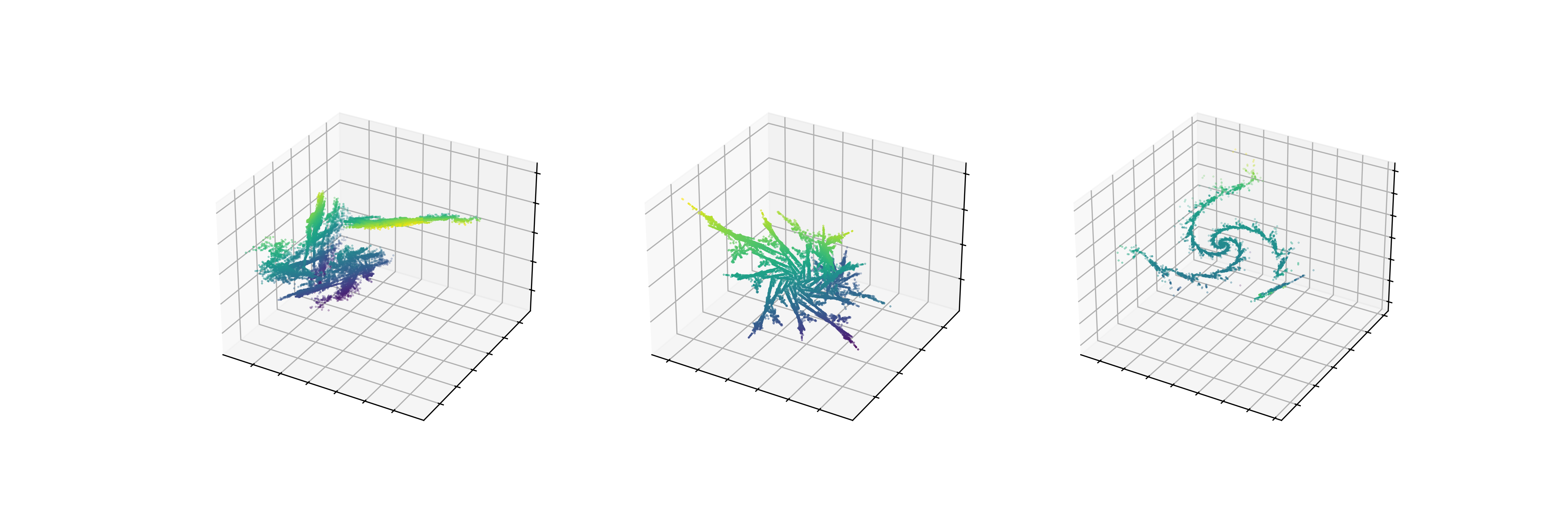}
        \caption{"Good" fractal class}
        \label{fig:good_fractals}
    \end{subfigure}%
    \begin{subfigure}{0.5\textwidth}
        \centering
        \includegraphics[width=\linewidth]{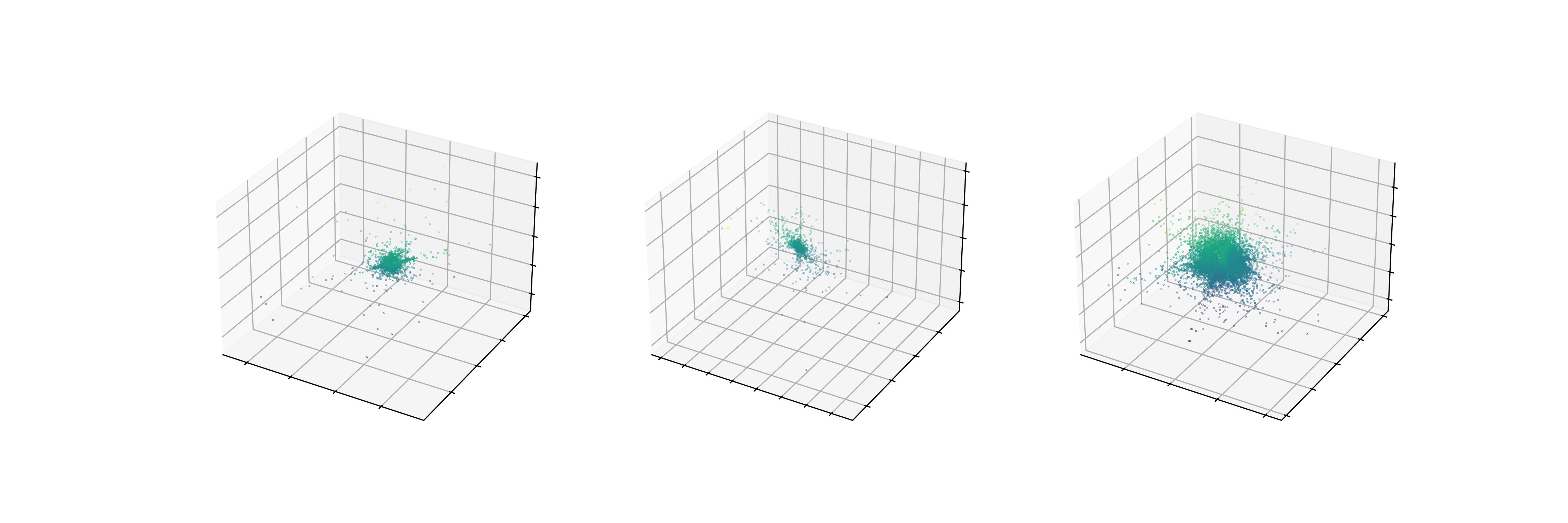}
        \caption{"Bad" fractal class}
        \label{fig:bad_fractals}
    \end{subfigure}

    \caption{Examples of manually annotated fractals in a 3D projection. In Figure~\ref{fig:good_fractals} we show complex and geometrically rich "Good" samples that exhibit self-similarity, while in Figure~\ref{fig:bad_fractals} we show collapsed or sparse "Bad" samples lacking structural detail and complexity.}
    \label{fig:manual_annotation}
\end{figure*}
\section{\uppercase{A Data-Driven Approach to 3D Fractal Quality}}

The cornerstone of fractal-based FDSL is the Iterated Function System: a method for constructing fractals using a finite set of contraction mappings on a complete metric space. IFS creates complex shapes by repeatedly applying a set of basic geometric transformations (like rotation, scaling, and translation) to an initial point, resulting in the fractal's characteristic self-similar structure. These transformations are represented by a matrix $\mathbf{A}$ and translation vector $\mathbf{b}$, whose parameters are randomly sampled.

However, not all IFS generated fractals observe complex self-similar structures that the fractals are best known for. Some functions push fractals to be extremely contractive, resulting in a collapsed blob, while others rapidly explode towards infinity leaving no complexity in shape. To determine which fractal sampling parameters are crucial for its final shape, data exploration was conducted.

\subsection{Manual Annotation}
To accurately find and extract features that guarantee geometrical complexity, fractals were randomly generated from a full parameter space (naive sampling), displayed in 3D and then manually annotated. Specifically, naive sampling here refers to drawing each value in the IFS's affine transformation matrices, $\mathbf{A}_i$, independently and uniformly from the range $U(-1, 1)$. In total, over $2000$ fractals were annotated into two categories: "Good" and "Bad". "Good" fractals exhibit complex geometry and are neither anisotropic, collapsed nor overly sparse. "Bad" fractals, on the other hand, appear collapsed in at least one dimension, form sparse blobs, or lack the structure and complexity that characterize the fractal's nature. Three examples of "Good" and "Bad" fractals are shown in Figure~\ref{fig:manual_annotation}. The annotations were made according to the following instruction:\\

    "\textit{The fractal must not be collapsed to a plane, line or a point. The fractal needs to display self-similarity. The fractal needs to have acceptable volume to differentiate multiple scales. The fractals must not appear as an ellipsoid without geometric complexity.}"\\

This instruction, while to some degree subjective, aided the process of sorting the fractals into the two classes in a systematic manner.

\subsection{Feature Extraction for Geometric Analysis}
In order to quantitatively analyze the generated fractals, we considered both statistical and topological features. Statistical features were used to capture per-system insights, while topological features were focused on insights based on the final fractal point cloud. 

The statistical features used are heavily focused on the the systems size of the IFS and the singular values of the generated matrices. This also includes features derived from the singular values such as the matrix determinant($|\det(\mathbf{A})| = \prod_{i=1}^3\sigma_i$) and the condition number ($\kappa=\frac{\sigma_1}{\sigma_3}$). Specifically, we consider the following features: 1) IFS System Size $N$, 2) the individual singular values ($\sigma_1$, $\sigma_2$, $\sigma_3$), 3) the condition number $\kappa$, 4) the matrix determinant $|\det(\mathbf{A})|$, and 5) difference between $\sigma_1$ and $\sigma_2$ or $\sigma_3$.

These features were chosen for their direct geometrical analogy. Each of the singular values describes how stretched or collapsed the resulting fractal is along that axis.  Minimal and maximal singular value are proportional to isotropy of the fractal; if their values differ significantly, or if the minimal singular values are small, the fractal will appear collapsed. Moreover, the ratio between maximal and minimal value is known as the condition number, a metric that shows how sensitive a matrix is to change. Absolute determinant of a system serves as a proxy to the volume distortion and it is derived from the product of singular values.

Since the system number $N$ varies per fractal, the statistical features were aggregated to form a single numerical value using either the mean or sum.

In addition to statistical measures, we also investigated topological features of the generated fractals, including the Euler Characteristics ($\chi$), Fractal Dimension ($D_B$), Top-K Eigenvalues, and Sphericity. These features aimed to quantify properties like connectivity, space-filling complexity, and dimensional anisotropy. However, during data exploration, we found that these topological features did not provide statistically significant insights for distinguishing between "Good" and "Bad" complex fractal geometries. Consequently, they were not utilized in the final feature selection or model training.

\subsection{Analysis of "Good" vs. "Bad" Geometries}

The goal of this analysis is to find a feature or a set of features that maximizes the class separation. Our starting point was a visual exploration of each feature's empirical density in both classes. Afterwards, the features were aggregated and plotted against each other to spot trends, correlations and class separability. Pair plots between the features were used as a direct way of assessing which feature combination could be used to produce geometrically good fractals.

To further validate the features, a Random Forest (RF) model was fitted to the extracted feature set to determine the importance each feature carries. Since the data is highly imbalanced as seen in Table \ref{tab:annotation_stats_distribution}, class balancing was utilized, features importance was permuted and 100 randomly initialized runs were conducted for feature importance stability.

\begin{table}[!t]
    \centering
    \caption{Manual annotation statistics for 2023 fractals, grouped by the number of Iterated Function System (IFS) functions ($N$). The table highlights the unbalanced class distribution due to prevalence of degenerate fractals.}
    \resizebox{\columnwidth}{!}{
        \begin{tabular}{lccccc}
            \toprule
            $N$ & Sampling Distribution (\%) & "Good" Class & "Bad" Class\\
            \midrule
            2 & $15.82\%$ & 63 & 257 \\
            3 & $13.40\%$ & 46 & 225 \\
            4 & $12.41\%$ & 28 & 223 \\
            5 & $19.92\%$ & 40 & 363 \\
            6 & $14.63\%$ & 31 & 265 \\
            7 & $11.81\%$ & 17 & 222 \\
            8 & $12.01\%$ & 19 & 224 \\
            \midrule
            Total& 100.00\% & 244 & 1779\\
            \bottomrule
        \end{tabular}
    }
    \label{tab:annotation_stats_distribution}
\end{table}

Table \ref{tab:feature_importance} shows the initial top 5 features sorted by importance according to the Random Forest (RF) model. To ensure that the final feature set selected for the quality filter provides independent discriminative power, we assessed feature redundancy. We implemented a correlation pruning strategy: starting with the most important feature, we iteratively select the next most important feature from the remaining pool. If the new feature exhibits an inter-correlation greater than $80\%$ with any feature already selected for the final set, it is discarded, and the next available feature in the ranked list is considered. This process continues until a final set of five highly important, yet sufficiently uncorrelated, features is established for the threshold quality filter.

\begin{figure}[!t]
    \centering
    \includegraphics[width=\linewidth]{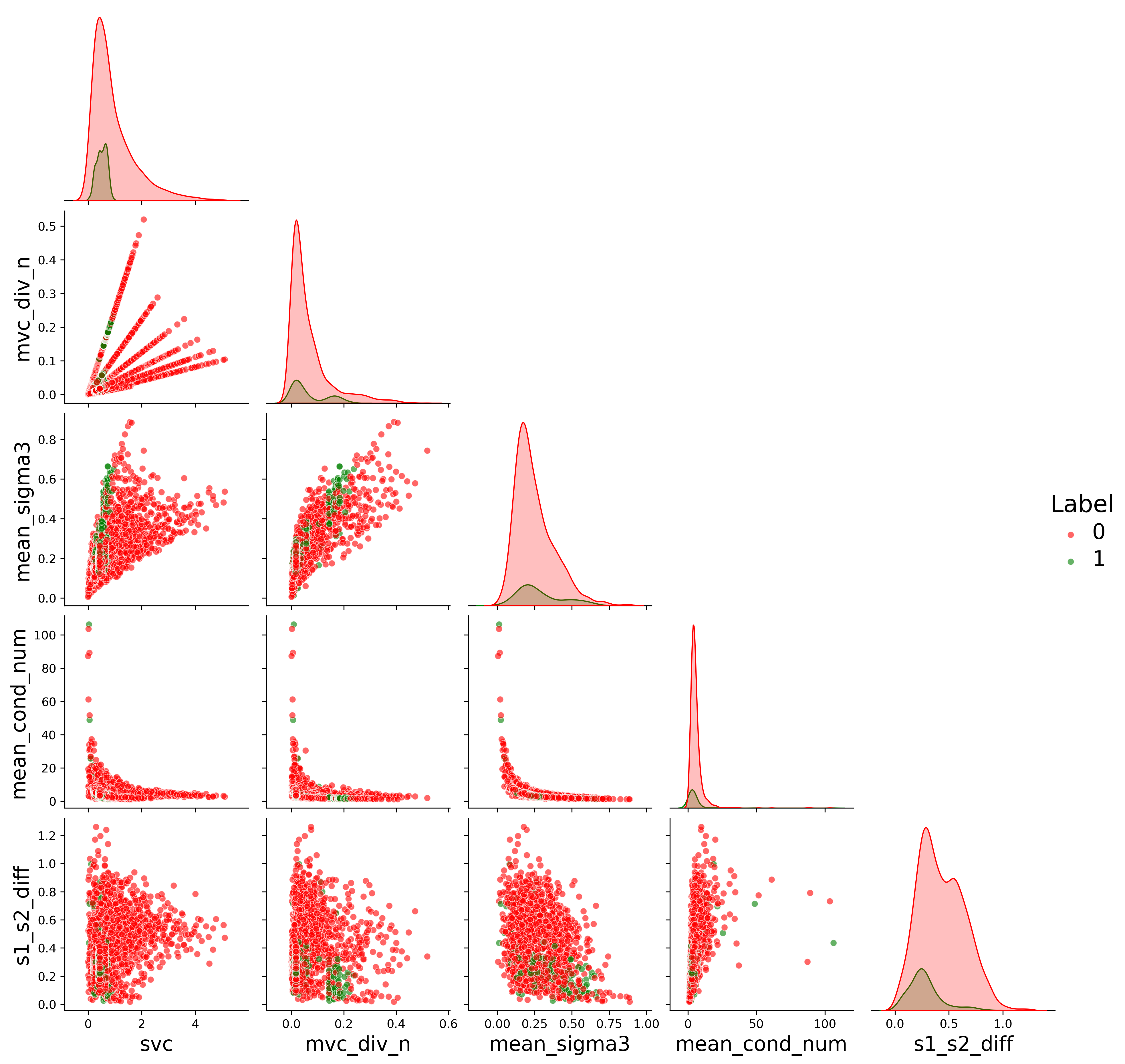}
    \caption{Pair plot of the top five features ranked by importance, showing only the lower triangle of the matrix. The diagonal panels present Kernel Density Estimates (KDEs) for each feature, highlighting the distributions of the "Good" (green) and "Bad" (red) classes.}
    \label{fig:pair_plot}
\end{figure}

Figure \ref{fig:pair_plot} presents the pair plot of the top five features. The visualization confirms that no combination of these features guarantees perfect separation of the "Good" class, although it remains useful for exploratory assessment of feature interactions.

\begin{figure}[!t]
    \centering
    \includegraphics[width=0.95\linewidth]{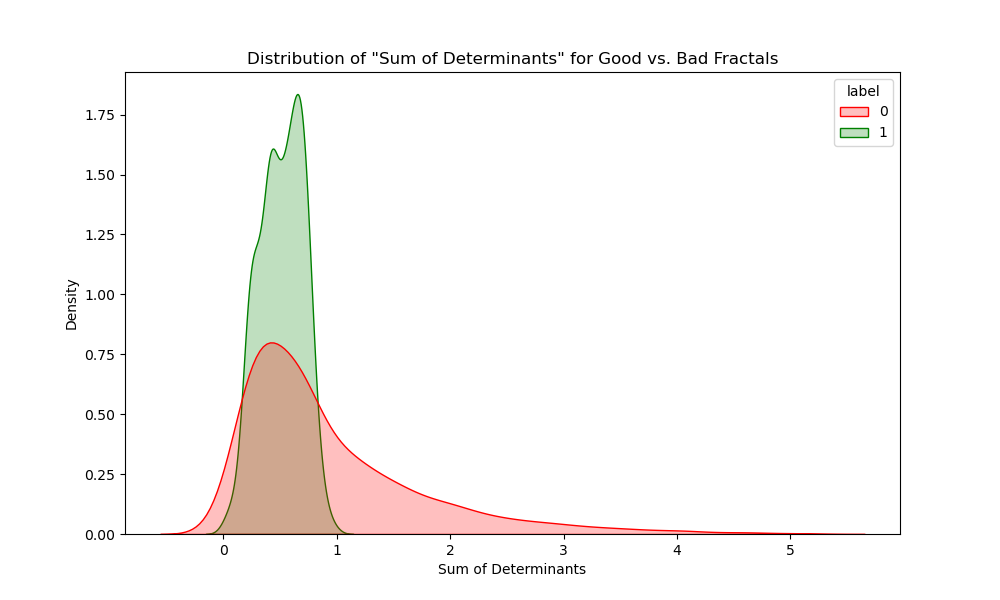}
    \caption{Kernel density estimate of the highest scoring feature by importance: Sum-of-$|\det(\mathbf{A})|$. The majority "Good" fractals are between $0$ and $1$ with the largest mode close to $1$. The "Bad" class peaks in a similar range, but it has a heavy tail toward higher values, which can be effectively filtered using a threshold.}
    \label{fig:svc_important}
\end{figure}

The Kernel Density Estimate (KDE) for the top-performing feature, Sum-of-Determinants, reveals interesting structural differences between the classes, as illustrated in Figure \ref{fig:svc_important}. While the modes (peaks) of the distributions are closely aligned, the distributions differ significantly in shape. The "Good" class exhibits a tighter, narrower distribution with a higher peak, indicating that complex fractals tend to cluster around a specific range of determinant values. Conversely, the "Bad" class is much wider and displays a pronounced heavy tail extending into higher values. This difference is crucial for a filtering strategy: by setting an upper threshold that clips this heavy tail, we can efficiently prune a large number of structurally defective (Bad) fractals. However, due to the overlap in the central distribution, this feature cannot provide perfect separation between the two classes alone, but it remains highly effective for filtering out degenerate and exploded fractals.

\begin{table}[!t]
    \centering
    \caption{Average feature importance values over 100 runs. Features with higher importance values contribute more significantly to classification performance.}
    \resizebox{\columnwidth}{!}{
        \begin{tabular}{lcc}
            \hline
            Feature & Average Importance \\
            \hline
            Sum-of-$|\det(\mathbf{A})|$ & 0.00802 \\
            Mean-of-$|\det(\mathbf{A})|$ / N & 0.00402 \\
            Mean ($\sigma_1 - \sigma_2$) & 0.00348 \\
            Mean-$\kappa$ & 0.00345 \\
            Mean $\sigma_3$ & 0.00260 \\
            \hline
        \end{tabular}
    }
    \label{tab:feature_importance}
\end{table}

\section{\uppercase{Methodology}}

Our methodology is divided into two parts. First, we describe the 3D Fractal Video Pipeline, a shared process used by all experiments to transform a static 3D point cloud into a dynamic video sequence. Second, we detail the four fractal generation strategies focusing on three classical approaches (Baseline, SVD-Control, RF-Filter) and our proposed Targeted Smart Filtering (TSF). The key difference between these strategies lies in how they sample and filter the initial IFS parameters ($\mathbf{A}, \mathbf{b}$) before the video pipeline is executed.

\subsection{3D Fractal Video Pipeline}

To generate a fractal point cloud, first the parameters of an Iterated Function System are sampled. Given a set of $N$ affine transformations $\mathcal{F} = \{{f_i}\ |\ i = 1, \ldots, N\}$, where $N$ is a uniformly sampled integer in $[2, 8]$, each $f_i$ is defined by a matrix $\mathbf{A}_i$ and a translation vector $\mathbf{b}_i$. Since the IFS is in 3D, a total of $12$ elements are sampled; $9$ for the $\mathbf{A}_i$ square matrix and $3$ for a three dimensional translation vector $\mathbf{b}_i$. For each affine transformation a selection probability, $p_i$, is assigned.

Once a valid set of IFS parameters the Chaos Game algorithm \citep{chaos_game} is applied to generate a point cloud $P$ of $m = 10\,000$ points by iteratively applying a randomly selected $f_i$ to the previous point. The selection probabilities $p_i$ are set proportional to $|\det(\mathbf{A}_i)|$ for more uniform coverage as shown by 
\citet{improving}.

To create a dynamic "action", the static point cloud $P$ is subjected to a sequence of geometric transformations. For each video, a set of target parameters is randomly sampled from rotation, translation, shear and a non-affine spatial warp. These parameters are then interpolated over $T$ frames, where $T$ is an integer uniformly sampled from $U(18,20)$, to create a temporally-coherent motion. To boost dynamics, temporal interpolation is stochastically selected to be either linear or sine ease-in-out; the following equation illustrates the sine ease-in-out function: 
\begin{equation}
    f(T) = \frac{1- \cos({\pi T)}}{2}
\end{equation}

This process ensures that each of the $C$ pre-training classes has a unique, consistent and dynamic motion profile.

\begin{figure*}[!t]
    \centering
    \includegraphics[width=\linewidth]{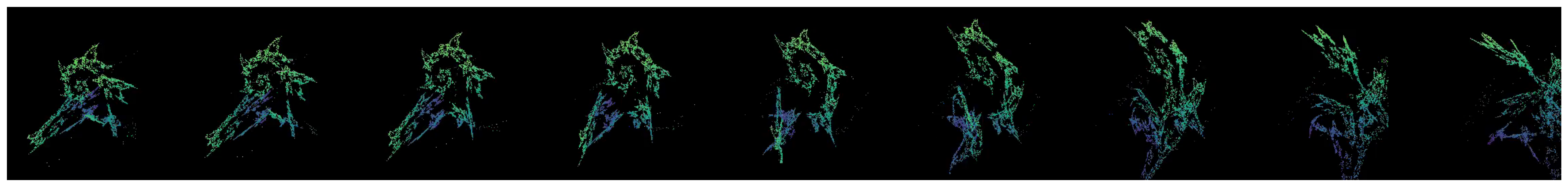}\hspace{1mm}
    \includegraphics[width=\linewidth]{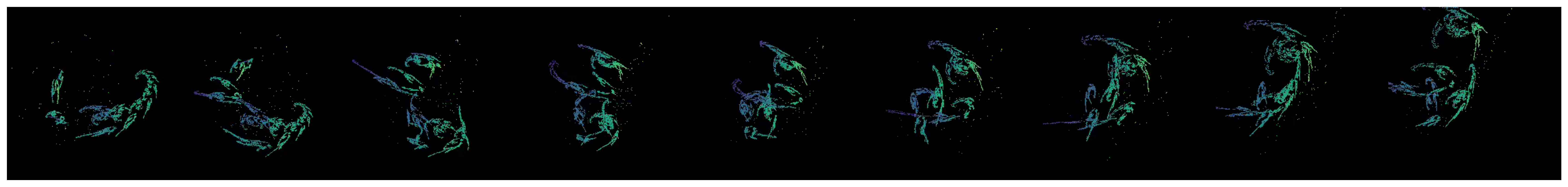}\hspace{1mm}
    \includegraphics[width=\linewidth]{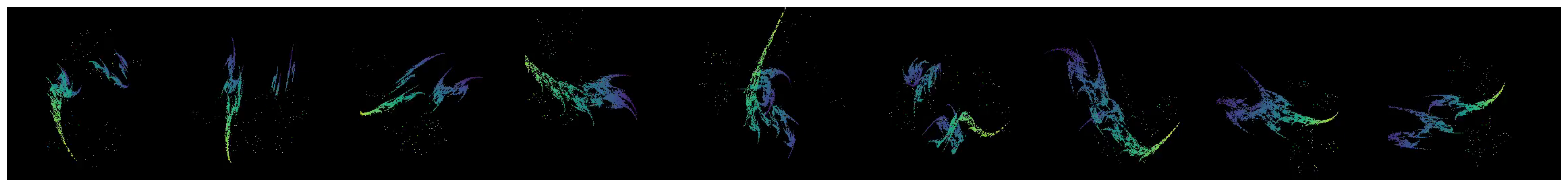}
    \caption{Three unrolled fractal transformation videos. Every other frame is rendered side-by-side to visualize how transformation affect fractal appearance over time.}
    \label{fig:unrolled}
\end{figure*}

Each of the $T$ transformed point cloud frames are rendered as a $256\times256$ RGB image from a fixed viewpoint using the Open3D library \citep{open3d}. Points are colored using the Viridis colormap based on their distance on the $z$-axis to provide consistent depth cue. The final sequence of $T$ frames is encoded as a 12 FPS MP4 video file, ready for pre-training. Example sequences are shown as unrolled frames in Figure \ref{fig:unrolled}.

\subsection{Fractal Generation Strategies}

The core of our investigation lies in the method used to sample the initial IFS parameters ($\mathbf{A}_i$, $\mathbf{b}_i$). The naive parameter space is vast and most random samples lead to degenerate geometries. Therefore, we compare four distinct strategies to find valid, non-degenerate fractal classes.

\subsubsection{Baseline: Naive Sampling + Variance Filter}

This strategy, adapted from the 3D point cloud work by \citet{mv-fractaldb}, serves as our primary baseline. The sampling process is as follows:

\begin{enumerate}
    \item Sample all 12 parameters for each of the $N$ systems from a uniform distribution, $U(-1,1)$.
    \item Run the computationally expensive Chaos Game to generate the full $10\,000$-point cloud $P$.
    \item Perform a post-hoc analysis on $P$, calculating the variance of the point cloud coordinates along the $x$, $y$ and $z$ axes.
    \item Apply a variance condition: The variance along each axis must be above a threshold (e.g., $0.05$). If the condition is not met, the sample is rejected as "degenerate" (e.g., collapsed to a plane or line), and the process restarts from Step 1.
\end{enumerate}

This method is simple but suffers from extreme computational inefficiency. The most expensive steps (generation and variance check) are performed before a sample can be rejected, leading to a very high cost per valid fractal. Qualitatively, the variance filter only guarantees that fractals will not be collapsed, but it lacks geometrical checks that rejects fractals that are blob-like, lacking structure or appearing as a shapeless ellipsoids.

\subsubsection{Method 1: SVD-Controlled Filter}

This method, inspired by the 2D work of \citet{improving}, avoids rejection sampling by constraining the parameter space from the start. Instead of sampling the matrix $\mathbf{A}_i$ directly, it is constructed by sampling its Singular Value Decomposed (SVD) components:
\begin{equation}
    \mathbf{A}_i = \mathbf{U}_i \Sigma_i \mathbf{V}_i^T
\end{equation}

The rotation matrices $\mathbf{U}_i$ and $\mathbf{V}_i$ are sampled by generating random Euler angles. Crucially, the singular values in the diagonal matrix $\Sigma_i$ are sampled uniformly from $U(0,1)$.

This process guarantees contractivity ($\sigma_{max} < 1.0$), resulting in a $0\%$ rejection rate and the fastest possible generation time (as shown in Table \ref{tab:generation_speed}). However, this method does not sample from the full, diverse $U(-1, 1)$ space. It enforces a strict geometric structure. As our results will demonstrate, this over-restriction is detrimental, as it removes the geometric diversity required for learning robust, transferable features. Note that the original Euler angle sampling of $\mathbf{U}_i$ and $\mathbf{V}_i$ has been tested against the quaternion angles in effort to avoid Gimbal lock \citep{gimbal_lock} and provide smoother interpolation but no apparent differences were found.

\subsubsection{Method 2: Data-Driven Investigation} 

After identifying the Baseline as inefficient and not warranting good geometry, and SVD-Control as "too restrictive", we conducted a deep, data-driven investigation to find a superior filter based on binary classification. This investigation, leveraging our $2023$ annotated fractals from Section 3, tested multiple hypotheses about geometric quality. These methods, while computationally expensive, were crucial for probing the nature of the parameter space. We tested four families of data-driven filters:

\begin{enumerate}
    \item Statistical Thresholding: The simplest data-driven approach. We used the analysis from Section 3.3 to build a filter from hard thresholds on the top-performing features. For instance, a constraint was implemented to accept only fractals where the Mean-of-$|\det(\mathbf{A})|$ fell within the range: $0.0276 <$ Mean-of-$|\det(\mathbf{A})| < 0.26$. These boundaries were determined by measuring the 5th and 95th percentiles (a 10\% symmetric trim) of the feature's empirical distribution across the annotated data. A more complex example involved conditioning the constraint on the system size. For each discrete value of $N$, a corresponding interquantile range for the Sum-of-$|\det(\mathbf{A})|$ was calculated, and newly generated samples were filtered based on both their $N$ and their Sum-of-$|\det(\mathbf{A})|$ value. Crucially, this method required post-hoc feature calculation for every new sample and, due to its rigid constraints, ultimately proved to be an excessively restrictive filtering mechanism.

    \item PCA-Based Variance Filters: We hypothesized that the Baseline's axis-aligned variance check was its key weakness. We developed an "upgraded" filter using Principal Component Analysis (PCA) to find the non-axis-aligned variance (i.e., the eigenvalues of the point cloud's covariance matrix). We tested two variants: 1) Naive + PCA Filter and 2) SVD-Control + PCA Filter. Results for both were below par. Both methods, while ideally filtering collapsed fractals, struggled to produce fractals with complex geometry, instead the fractals appeared as sparse blobs.
    \item The Failed Search for a 3D $\alpha$-factor: We attempted to replicate the 2D success of \citet{improving} by finding a 3D $\alpha$-factor (a simple linear quality score). We trained a Linear SVM on our $2023$ annotated samples, using the mean singular values ($\bar\sigma_1$, $\bar\sigma_2$, $\bar\sigma_3$) as input features to find a separating hyperplane. This attempt failed. Unlike in 2D, we found that no simple linear combination of mean singular values could reliably separate "Good" from "Bad" 3D fractals, even with the larger annotated data collection. This suggests the geometric complexity of 3D IFS is non-linear and cannot be captured by such a simple metric.

    \item Classifier-as-Filter (RF-Filter): A more complex filter was implemented using the trained Random Forest model itself. A fractal candidate was generated using the Baseline sampling, its features were extracted and passed to the RF classifier. The fractal is only kept if the classifier predicts it as "Good" (e.g., \texttt{predict\_proba(Good) > 0.5}). This approach was computationally expensive due to feature extraction and model inference compared to other methods. While capturing well-formed fractals, its performance is one of the worst since the passed fractals are very similar in appearance.

\end{enumerate}

This exhaustive, data-driven investigation led to a critical insight: all filters based on quality (human-perceived, statistical, or classifier-based) were both computationally expensive and detrimental to performance. They all over-restricted the data. This proved that we needed a filter that was 1) fast (pre-generation), 2) un-opinionated about quality, and 3) focused only on removing degeneracy. This insight led directly to the development of TSF.

\subsubsection{Method 3: Our Proposed Targeted Smart Filter (TSF)}

Our proposed Targeted Smart Filter (TSF) is the solution derived from the failures of the other methods. It is designed to find the optimal "sweet spot", achieving the efficiency of SVD-based checks without the restrictive nature of SVD-component-sampling, all while retaining the diversity of the Baseline's naive sampling. TSF is a lightweight, two-stage pre-generation filter that operates directly on the naively sampled $U(-1,1)$ parameters, before the expensive Chaos Game is run:

Stage 1: Contractivity Filter. For each of the $N$ matrices $\mathbf{A}_i$ in the naively sampled IFS, we compute its largest singular value, $\sigma_{max}(\mathbf{A}_i)$, by computing the eigendecomposition of the matrix $\mathbf{A}_i$ (also known as the spectral norm $||\mathbf{A}||_2$). This is the exact mathematical measure of contractivity. If any system in the IFS has $\sigma_{max}(\mathbf{A}_i) \geq 1.0$ , the entire IFS is immediately rejected. This single check efficiently prunes the vast majority of diverging (exploding) samples.

Stage 2: Collapse Filter. For the candidates that pass Stage 1, we then compute the smallest singular value, $\sigma_{min}(\mathbf{A}_i)$, for all $N$ matrices. If any $A_i$ has $\sigma_{min}(\mathbf{A}_i) < \epsilon $ (where $\epsilon$ is a small threshold, e.g., $0.2$), the entire IFS is rejected. This efficiently filters out transformations that would collapse the 3D point cloud onto a 2D plane or a 1D line.\\

The crucial change in TSF opposed to SVD-Control is that samples are taken from the full, naive $U(-1,1)$ parameter space and SVD properties are used as a fast, lightweight filter.

This hybrid approach allows TSF to preserve the rich, complex, and diverse geometry of the naive sampling space while also being computationally efficient. As our results will demonstrate, this combination provides the optimal trade-off, leading to the second fastest valid-fractal generation and the best-performing downstream model.

\section{\uppercase{Experimental Setup}}

The experimental phase of this work is structured to systematically compare the four proposed fractal generation methods (Baseline, SVD-Control, RF-Filter, and TSF) on their ability to produce pre-training data that yields high performance on downstream action recognition tasks. This section details the datasets used, the model architecture selected, the training protocols for both pre-training and fine-tuning, and the metrics utilized for evaluation.

\subsection{Datasets}
 
\textbf{Pre-Training Datasets:} To facilitate a direct, fair comparison between generation strategies, four primary pre-training datasets were constructed, one for each method: Baseline, SVD-Control, RF-Filter, and our proposed Targeted Smart Filtering (TSF). Each of the four datasets consists of $500$ unique fractal classes. For each class, $100$ video instances were generated using the 3D Fractal Video Pipeline (Section 4.1), resulting in $50\,000$ training videos. A separate validation set of $5\,000$ videos ($10\%$ instances per class) was generated to monitor generalization during pre-training.

\noindent\textbf{Downstream Action Recognition Datasets:} We evaluate transfer performance on two standard video action recognition benchmarks, following their official three training/test splits and reporting the mean accuracies:
\begin{itemize}
    \item UCF101 \citep{ucf101}: $13\,320$ videos across $101$ action categories.
    \item HMDB51 \citep{hmdb51}: $6\,766$ videos across $51$ challenging action categories.
\end{itemize}

\noindent These datasets were chosen as they are commonly used datasets within the broader action recognition domain.

\subsection{Model Architecture and Training Protocol}
\textbf{Model Architecture:} 
Following existing FDSL work, we chose the ResNet-50 backbone integrated with the Temporal Shift Module (TSM) \citep{tsm}. This provides an efficient architecture capable of robust spatio-temporal modeling at a 2D computational cost, suitable for extensive comparative studies. This model was also chosen as to allows us to directly compare with the results achieved by \cite{2dfractals_ar}, who used interpolation between two 2D fractal to generate a video. 

\noindent\textbf{Training Protocols:} During pre-training the model was randomly initialized, trained for 25 epochs using the AdamW optimizer \citep{adamw} with cosine annealing schedule. The initial learning rate was set to $10^{-3}$. The batch size was $128$. For fine-tuning we initialized with the pre-trained model and use the same hyperparameters across both downstream datasets. We fine-tune for $100$ epochs using the SGD optimizer, as well as a standard learning rate scheduler starting at $10^{-3}$. The learning rate was reduced by $10$ at $50\%$ and $75\%$ of total epochs. The batch size was set to $64$. 

\subsection{Augmentation and Evaluation Metrics}
All video inputs were downscaled to $112\times 112$ pixels and clipped to a length of $8$ frames, which were randomly sampled from the sequence. Next, the pre-training augmentations were applied to encourage robust feature learning. Augmentations were directly mirrored from the \citet{2dfractals_ar} and include rotation, grouping, camera shake, random zoom, blur, etc.

For downstream performance, we report Top-1 and Top-5 train and validation accuracy. For generation efficiency, we report the Mean Wall-Clock Time (seconds) and the Mean Rejection Rate (\%) required to generate $100$ valid classes, averaged over $10$ runs. \\

We primarily compare the fractal sampling methods among each other. We also include a comparison against various baselines and comparative methods: Training from scratch, a 2D fractal video dataset generated as described by \cite{2dfractals_ar}, and the pre-training with the Kinetics-400 dataset. It should be the noted that the Kinetics baseline is not an equal comparison due to the Kinetics dataset having four times larger area per frame ($224\times224$ vs $112\times112$) and five times larger dataset ($250\,000$ vs $50\,000$ instances). The Kinetics baseline is also pre-trained for $100$ epochs opposed to our $25$ epochs.

\section{\uppercase{Results and Discussion}}

\begin{table*}[!t]
    \centering

    \caption{Downstream Top-1 and Top-5 Validation Accuracy (\%) on HMDB51 and UCF101. The "Data Type" column denotes which kind of pre-training data was used: Natural images (Nat), 2D Fractals (2D), or 3D Fractals (3D).}

    \label{tab:train-results}

    \begin{tabular}{lccccc}
        \toprule

        Pre-training Method & Data Type & \multicolumn{2}{c}{HMDB51} & \multicolumn{2}{c}{UCF101} \\

        \cmidrule(lr){3-4} \cmidrule(lr){5-6}

         && Top-1 & Top-5 & Top-1 & Top-5 \\
        \midrule

        From Scratch & - & 31.5 & - & 70.3 & - \\
        \citet{2dfractals_ar} & 2D & 56.5 & - & 81.8 & - \\
        Kinetics \citep{kinetics} & Nat & 70.1 & - & 95.3 & - \\
        
        \midrule

        Baseline (Naive + Var.) & 3D & 49.3 & 79.4 & 77.7 & 95.1 \\
        SVD-Control Filter & 3D & 47.4 & 78.3 & 75.7 & 94.3 \\
        Data-Driven (RF-Filter) & 3D & 44.3 & 73.5 & 74.4 & 92.8 \\

        Targeted Smart Filtering (TSF) & 3D & 48.5 & 79.5 & 78.3 & 95.4 \\
        
        \bottomrule
    \end{tabular}
\end{table*}

\begin{table}[!t]
    \centering
    \caption{Average time required to generate parameters for $100$ fractal classes and corresponding number of rejected samples (failed attempts). The proposed Targeted Smart Filter (TSF) method achieves the best parameter space exploration in second shortest time, triumphing over naive method substantially.}
    \resizebox{\columnwidth}{!}{
        \begin{tabular}{lcc}
            \hline
            Method & Mean Time (s) & Mean Rejections \\
            \hline
            Baseline (Naive + Var) & 152.68 & 6974 \\
            SVD-Control & 0.69 & 0 \\
            Data-Driven (RF-Filter) & 27.17 & 1852 \\
            TSF (Ours) & 1.54 & 28860 \\
            \hline
        \end{tabular}
    }
    \label{tab:generation_speed}
\end{table}

We present our results on HMDB51 and UCF101 in Table \ref{tab:train-results}, and observed that the 3D fractal pre-training as a whole is effective, outperforming from scratch training, though still falling short of other pre-training setups. Specifically, we find that all methods confidently outperform training from scratch ($+17\%$ on HMDB51, $+8\%$ on UCF101). This proves the method is viable and beneficial to use as a pre-training mean in FDSL.

We also present results in terms of fractal generation efficiency in Table \ref{tab:generation_speed}, determined by measuring the time required to produce parameters for $100$ classes. A total of $10$ runs per method were conducted to ensure stability. The performance test was executed on an i7-13700H CPU. We find that TSF is approximately 100 times faster than the Baseline and RF-Filter. It is also clear that the Naive Baseline is extremely inefficient, requiring over $150$ seconds and only rejecting $6974$ samples to find $100$ valid classes. The SVD-Control is the fastest method that by definition does not generate rejectable samples, however its poor downstream performance seen in Table \ref{tab:train-results} is the limiting factor, as is the similar case with RF-Filter. 

These observations makes TSF the clear choice: it is the only method that is both computationally efficient and provides the best downstream performance. Due to its fast sampling speed, the TSF method is a promising candidate for on-the-fly dataset generation during training, effectively becoming a negligible computational overhead. 

A crucial finding is how the fractals are filtered matters significantly. Both the SVD-Control method and the complex Data-Driven (RF-Filter) method perform worse than the simple Naive + Variance baseline. The SVD method, while fast, is too restrictive. It strips necessary variance of the fractals that instill robust model features that are beneficial for downstream tasks. The RF-filter, trained on the annotated data, also filters too aggressively which is reflected on even worse performance. Simple inference probability decrease, to allow for more fractals to be classified as good, does not add any value because it only allows for more degenerate fractals to appear rather than increasing diversity. This confirms that over-filtering and restricting the parameter space harshly is detrimental. 

The novel Targeted Smart Filter (TSF) method was constructed to address the earlier pitfalls. It is the only method to outperform the naive baseline, achieving the highest accuracies. This demonstrates that TSF successfully finds the optimal restriction to the full parameter space. It is more effective than a simple variance check, but not restrictive of the geometric diversity which SVD and RF-Filter discard.

\begin{figure*}[!t]
    \centering
    \includegraphics[width=0.95\linewidth, height=3.5in]{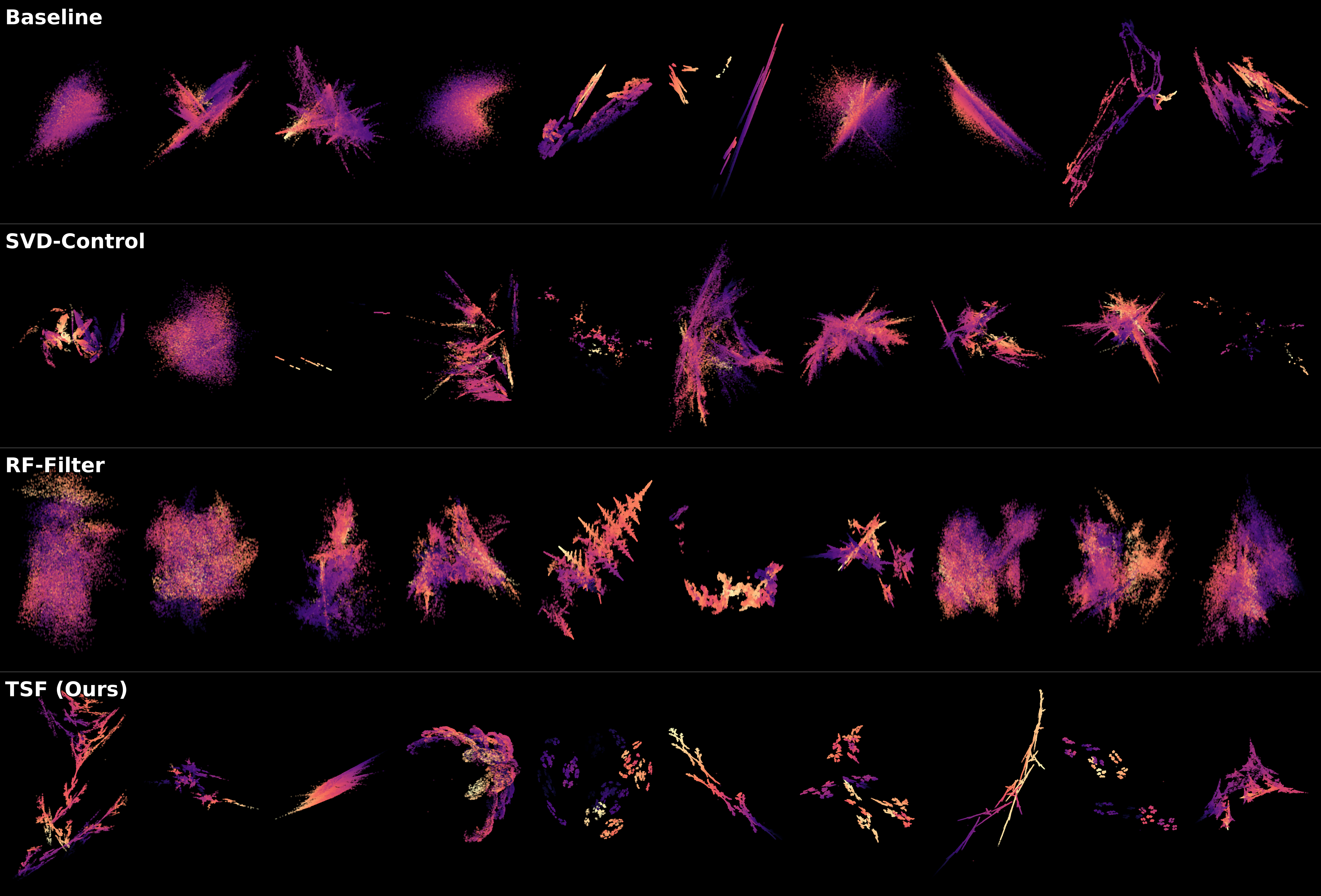}
    \caption{Qualitative analysis of generated fractals using different sampling method. All fractals are generated with $20\,000$ points and colored with Magma colormap along their $z$-axis. All other methods exhibit similar shortcomings, except TSF. Some fractals are dense with no complex geometry or self-similarity. They resemble simple point cloud blobs. The TSF method proves quality and geometrical complexity. It is not sparse, nor unstructured, it has sharp contours and diverse fractals.}
    \label{fig:qualitative_inspection}
\end{figure*}

These observations are clearly depicted in Figure \ref{fig:qualitative_inspection}. SVD-Control produces structurally similar fractals due to its restricted sampling space, which explains its poor transferability. Similarly, the RF-Filter and Baseline exhibit similar tendencies of either sacrificing complexity (RF-Filter) or including too many degenerate samples (Baseline). Conversely, TSF output is highly diverse, exhibiting the complex, self-similar geometry and sharp contours required to train robust feature detectors. The importance of contours is confirmed by Kataoka et al. in their auto-generating contours study \citep{contours}. Therefore, we attribute the success of TSF to its ability to preserve the inherent complexity of the full parameter space while only eliminating geometric degeneracy.

Lastly, in order to evaluate the robustness and scalability of our proposed TSF method, we conduct an ablation study on the pre-training dataset size. We compare TSF against the Baseline (Naive+Var) to determine if the performance gains are consistent as the dataset scales. We generated larger datasets of $100\,000$ and $200\,000$ instances (while preserving the $500$ class count) and report the fine-tuning performance in Table \ref{tab:ablation_size}.

\begin{table}[!t]
\centering

\caption{Ablation Study on Dataset Size. Performance (Top-1 \%) is shown for the Baseline and our proposed TSF method as the number of pre-training instances increases ($500$ classes constant). TSF consistently scales better than the Baseline.}
\label{tab:ablation_size}
\resizebox{\columnwidth}{!}{

\begin{tabular}{l l c c} 
\toprule

Method & Size & HMDB51 Top-1 & UCF101 Top-1 \\
\midrule
Baseline (Naive + Var) & 50k & 49.3 & 77.7 \\
                       & 100k & 49.0 & 78.4 \\
                       & 200k & 50.8 & 80.4 \\
\midrule

TSF (Ours)             & 50k & 48.5 & 78.3 \\
                       & 100k & 51.8 & 81.5 \\
                       & 200k & 54.6 & 82.5 \\
\bottomrule
\end{tabular}
}
\end{table}

The results in Table \ref{tab:ablation_size} show a clear and positive trend: performance on both datasets generally scales favorably with the number of pre-training instances. It is worth noting that the Baseline method shows a minor performance dip on HMDB51 at 100k instances (from $49.3\%$ to $49.0\%$), which we attribute to normal training variance. However, the overall scaling trend is upward for both methods.

More importantly, the rate of scaling demonstrates the superiority of our TSF method. Increasing the TSF dataset from 50k to 200k instances yields a significant performance gain of $6.1\%$ on HMDB51 (from $48.5\%$ to $54.6\%$) and $4.2\%$ on UCF101 (from $78.3\%$ to $82.5\%$). In contrast, the Baseline method exhibits much poorer scaling, gaining only $1.5\%$ on HMDB51 and $2.7\%$ on UCF101 over the same data increase. This demonstrates that our TSF method is not only highly efficient but also scalable. It confirms that the performance gains are robust and that further improvements can likely be achieved by training on even larger TSF-generated datasets.

\section{LIMITATIONS}
In this work we have demonstrated the feasibility of pre-training action recognition networks on 3D fractals, as well as how to efficiently sample good 3D fractals. 
However, our work is not without limitations. The use of 3D fractals is so far lacking behind simply interpolating between two 2D fractals \cite{2dfractals_ar}. In order to understand why this is we believe attention should be placed on selecting appropriate video transformations. 
Secondly, our filtering experiments and evaluation of TSF were limited to 3D fractals and the action recognition task. We believe that the demonstrated efficiency and performance of TSF warrants further analysis on on both 2D fractals and other tasks.

\section{CONCLUSIONS}

In this paper we have presented a comprehensive pipeline for generating 3D fractal-based videos for action recognition pre-training. It systematically compared different generation strategies, from a simple variance filter to a novel efficient heuristic filter. Our core findings are:
\begin{enumerate}
    \item Pre-training on synthetic 3D fractal videos significantly outperforms training from scratch, demonstrating the viability of 3D FDSL for video.
    \item Overly-restrictive filtering, whether through rigid geometric constraints (SVD-Control) or complex human-in-the-loop classifiers (RF-Filter), is detrimental to performance, as it removes critical geometric diversity. 
    \item A novel Targeted Smart Filtering (TSF) was proposed that achieves the optimal trade-off between parameter space exploration and geometric degeneracy filtering. It efficiently explores the full parameter space leading to new top performance for 3D fractal pre-training, while also being significantly faster to generate than the naive baseline.
\end{enumerate}

\noindent This work demonstrates that the most effective FDSL datasets are not necessarily those that are most aesthetically pleasing, but those that are most efficiently and intelligently diversified. We provide this with our TSF method, a practical and superior path to generating such datasets compared to prior filtering methods.

\section*{\uppercase{Acknowledgments}} This work was supported by the Pioneer Centre for AI (DNRF grant number P1).

\bibliographystyle{apalike}
{\small
\bibliography{example}}

\clearpage
\end{document}